\documentclass[letterpaper, 10 pt, conference]{ieeeconf}
\IEEEoverridecommandlockouts
\usepackage{cite}
\usepackage{amsmath,amssymb,amsfonts}
\usepackage{algorithmic}
\usepackage{graphicx} 
\usepackage{color}
\usepackage{xcolor}
\usepackage{subcaption}
\usepackage{hyperref}
\usepackage{float}
\usepackage{diagbox}
\usepackage{comment}

\def\BibTeX{{\rm B\kern-.05em{\sc i\kern-.025em b}\kern-.08em
    T\kern-.1667em\lower.7ex\hbox{E}\kern-.125emX}}
\begin{document}

%\title{\LARGE \bf Mission Planning for Conditionally Complex Autonomous Navigation using ChatGPT and ROS2}

\title{\LARGE \bf Leveraging LLMs for Mission Planning in Precision Agriculture }

\author{Marcos Abel Zuzu\'{a}rregui \qquad Stefano Carpin%
\thanks{
The authors are with the Department of Computer Science and Engineering, University of California, Merced, CA, USA.
This work is partially supported by the IoT4Ag Engineering Research Center funded by the National Science Foundation (NSF) under NSF Cooperative Agreement Number EEC-1941529 and under grant CMMI-2326310. Any opinions, findings, conclusions, or recommendations expressed in this publication are those of the author(s) and do not necessarily reflect the view of the National Science Foundation.}% <-this %
}

%\author{\IEEEauthorblockN{Marcos Zuzu\'{a}rregui} 
%\IEEEauthorblockA{\textit{University of California, Merced} \\
%\textit{Department of Electrical Engineering and Computer Science}\\
%Merced, California, USA \\
%mzuzuarregui@ucmerced.edu}
%}

\maketitle

\begin{abstract} 
Robotics and artificial intelligence hold significant potential for  advancing precision agriculture. While robotic systems have been  successfully deployed for various tasks, adapting them to perform diverse  missions remains challenging, particularly because end users often lack  technical expertise. In this paper, we present an end-to-end system that  leverages large language models (LLMs), specifically ChatGPT, to enable  users to assign complex data collection tasks to autonomous robots using  natural language instructions. To enhance reusability, mission plans are  encoded using an existing IEEE task specification standard, and are executed  on robots via ROS2 nodes that bridge high-level mission descriptions with  existing ROS libraries. Through extensive experiments, we highlight the  strengths and limitations of LLMs in this context, particularly regarding  spatial reasoning and solving complex routing challenges, and show how our proposed implementation overcomes them.
\end{abstract}

%\begin{IEEEkeywords}
%    graph neural network, orienteering, message passing, MCTS, stochastic, chance constraint
%\end{IEEEkeywords}

%\input{sections/introduction}

\section{Introduction}
Mission planning (MP) can be defined as ``any system that plans the operations of another system or any of its components'' \cite{jones_review_1993}
and is central to the deployment and adoption of autonomous robotic
systems operating in semi-structured or unstructured environments
 \cite{kalluraya_multi-robot_2023, askarpour_mind_2020,brumitt_dynamic_1996}.
 MP is related to other classic areas in robotics and planning, such 
 as task and motion planning \cite{itamp} and mission specification 
 through linear temporal logic \cite{ltl}. %Boundaries between 
 %these different areas are often blurred and up to subjective
 %interpretation. 
 Solving MP problems involves multiple challenges.
One deals with synthesizing a plan leading to the  desired outcome.
In classic AI research dealing with assembly tasks, it is often assumed
that the outcome of actions is predictable, and notwithstanding the
problem remains often very difficult due to the large branching factor
in the search space. Another aspect relates to \emph{how} one informs
the MP system about the desired outcome, especially when 
the desired  behavior  involves complex sequences of actions
with uncertain outcomes.
\begin{figure}[htb]
\centering
\includegraphics[width=\linewidth]{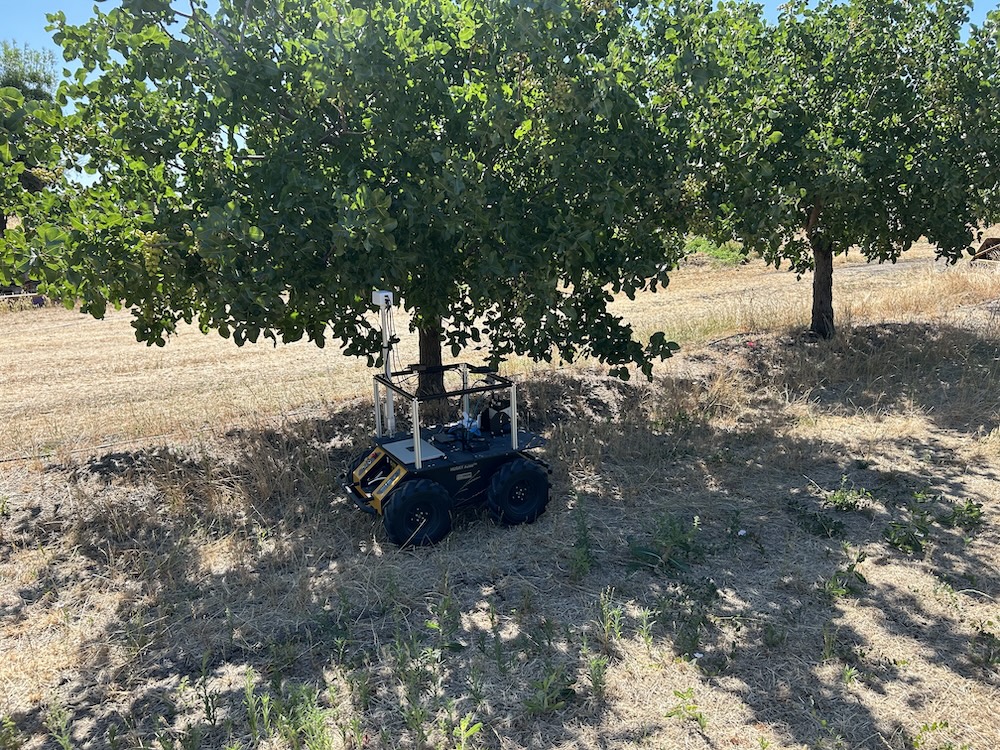}
\caption{A task in precision agriculture: a robot acquiring thermal imaging
of a pistachio tree for water stress detection.}
\label{fig:graph20_ucm}
\vspace{-5mm}
\end{figure}
Our ongoing research in robotics for precision 
agriculture \cite{CarpinRAM2023} motivates our  interest
for MP in this application area, and there are various
interrelated challenges we tackle in this paper. As
agricultural robots are expected to be operated by non specialists,
a natural requirement is to empower users with the ability of 
synthesizing complex mission plans without having to deal
with low level details or understanding system level complexities.
Moreover, in precision agriculture missions, robots operate in semi-structured
environments where the outcome of actions is unpredictable. 
Importantly, the outcome of the mission 
 may depend on data collected at run time. For example, a robot may be
tasked with collecting pictures of some trees (see e.g.  Figure \ref{fig:graph20_ucm}), and the actions
may be adjusted at run time based on partial results (e.g., if trees
appear to be stressed, the robot may be additionally required to 
collect soil moisture samples). Another layer of complexity arises from the fact that robots operating in rural areas often lack network connectivity, making it impossible to access remote services in real time. Additionally, a critical component of these applications is the need for spatial planning and reasoning, both in quantitative terms (e.g., taking three moisture measurements 15 meters apart) and qualitative terms (e.g., measuring soil nitrate levels on the east side of the orchard).

Starting from these premises,  we present an end-to-end system that tackles these challenges exploiting the novel abilities offered by large language models (LLMs).
With the advent LLMs, the intersection between natural language processing, planning and robotics is seeing tremendous growth \cite{mower_ros-llm_2024, kannan_smart-llm_2024}, in an effort to make human-robot interactions seamless. 
% LLMs, particularly OpenAI's ChatGPT, have been the focus of various research projects about their potential and utility \cite{ray_chatgpt_2023, wen_future_2023}. 
% Their success in simplifying every day software related tasks, GitHub CoPilot, to helping students with their homework has seen an increase in productivity across the tech world {\color{red} cite?}. 
% In robotics, they are being leveraged as a tool for MP and writing control logic of the high and low order {\color{red} cite}. 
% However, much of the attention is focused on manipulation logic design using LLMs. 
Considering our focus on precision agriculture, we question whether the same 
principles applied in large language models (LLMs) could be relevant in this 
domain. For instance, when planning a robot's path through an orchard, 
numerous contingencies must be taken into account, which can influence the 
trajectory. While the primary goals are being met, it is also essential to 
optimize resources and adhere to various constraints.
%Whether specific to the mission or uncovered while carrying out the mission, this variability is critical in a successful trial. 
%Now, we ask, 
\textit{Can current LLM models solve  mission planning problems in precision
agriculture requiring to handle uncertain outcomes, spatial awareness and resource optimization?}

% Specifically, commercial products that have come out not only require a steep up front investment but then require time and effort, which - ultimately - ends up being more money. 
% Another common problem in these applications is the limited connectivity of robots in and around a farm. 
% Due to their isolated nature, communication is difficult, which makes keeping a cloud based planner in the loop difficult.

% One of the problems MP aims to solve for in robotics is the stochastic orienteering problem (SOP): an APX-hard optimization problem \cite{doi:10.1137/050645464}. {\color {red} can I site my own unpublished work?} In agriculture, we see this applied as data collection in orchards, harvest assistance, weed spraying, among other applications {\color{red} cite}. 
% % The SOP is a variant of the traveling salesman problem (TSP) in that each vertex of a graph has some arbitrary value but the solver has a limited amount of budget to travel. In applied versions of these NP-hard problems, a robot will not know its future state until after it has carried out an action, due to stochasticity of the real world. 
% We intend to display our LLM MP solution with respect to the data collection problem in real world graphs such as \ref{fig:graph20_ucm}. 

In this paper,  we examine the use of LLMs to generate mission plans and explore 
the standardization of LLM-generated mission plan outputs based on the 
recently published IEEE Standard for task specification. \cite{noauthor_ieee_2024}. 
We will also show the limitations of ChatGPT and the necessity 
to augment a mission plan with human designed components to solve 
mission relevant resource optimization problems.
The contribution of this paper are the following:
\begin{itemize}
    \item we present an LLM to robotic task execution pipeline for autonomous  navigation and data collection; 
    \item we show how our use of a LLM mission planner can be 
    integrated with the IEEE standard 1872.1-2024 \cite{noauthor_ieee_2024};
    \item we investigate whether mission plans generated leveraging
    LLMs can effectively reason about space and consider stochastic
    optimization constraints typical of the precision agriculture domain;
    \item we validate our proposed system in the field and show limits 
    and strengths.
\end{itemize}

The rest of the paper is  as follows. 
Selected related work is presented in Section \ref{sec:sota}. 
In Section \ref{sec:method} where we describe the system we developed,
experiments detailing our findings are given in Section \ref{sec:results}, and conclusions are given in Section \ref{sec:conclusions}.

\vspace{-3mm}
\section{Related Literature} \label{sec:sota}
\subsubsection{Mission Planning}
Literature in MP for robotics is vast, and
 over the years  we have seen an increase in mission planning literature ranging from distributed and dynamic mission planning \cite{brumitt_dynamic_1996, kalluraya_multi-robot_2023} to applied mission planning \cite{askarpour_mind_2020} to the ever popular manipulator planning \cite{huang_inner_2022, ahn_as_2022, huang_language_2022}. While often mission plans are specified using linear temporal logic (LTL) \cite{kalluraya_multi-robot_2023, kalluraya_resilient_2023, kumar_linear_2016}, this creates a difficult interface for end users. Notably, many of these papers focus on some form of runnable code generation. We differ by implementing a well-formed, decomposed task representation by way of \cite{noauthor_ieee_2024} to standardize our mission plan output. For example, \cite{rizk_cooperative_2019, mower_ros-llm_2024, kannan_smart-llm_2024} used logical, non-standardized task decomposition definitions generated in Python. To our knowledge, this paper is the first to use this type of generic non-executable framework in mission planning. 

\subsubsection{LLMs}
While solving MP problems, many authors also integrate modern LLMs into their architectures, such as  \cite{mower_ros-llm_2024, kannan_smart-llm_2024, liang_code_2023, huang_language_2022}, among others.
However, in all of these papers and many others, the task is typically some form of robotic manipulation. 
While there exists literature in autonomous navigation \cite{elhafsi_semantic_2023, xu_drivegpt4_2024, chen_driving_2024} using LLMs, to the best of our knowledge, we have yet to read anything about precision agriculture mission planning using LLMs. 
The key difference in these autonomous navigational papers is that they 
mostly focus on semantic reasoning, while we address task representation.

Where previously LTL \cite{jansen_can_2023} and Planning Domain Definition Language (PDDL) \cite{noauthor_ieee_2024} were used extensively in the past to generate mission plans, the difference in our experiments is that we use ChatGPT to generate Extensible Markup Language (XML) against an XML Schema Definition (XSD). 
While seen before in \cite{tam_let_2024} where LLM output is well-formed by format-restriction prompting, our contribution also incorporates a rationale behind the format through \cite{noauthor_ieee_2024}.
% While many of the above LLM papers embody the same concept with respect to task representation, the most notable difference is that instead of designing our own robotic task framework, we leverage \cite{noauthor_ieee_2024}. 

% \subsection{Stochastic Orienteering}
% {\color{red} do we port what we had from the other paper? can we do that?}
% Generally speaking, deterministic orienteering or the traveling salesman problem has seen much exploration and many implementations have near-optimal solutions \cite{liu_boosting_2021, DBLP:journals/corr/DaiKZDS17, 9109309}. However, the larger open question in this variant of NP-hard problems is the stochastic variant, particularly that with a chance constraint. While \cite{9636669, 9636104, 9926510} have explored this problem in depth, the solution is based on heuristic and performance is on the order of seconds. In our implementation, which leverages the latest in machine learning - a graph neural network (GNN) - we see real-time performance on the order of hundreds of milliseconds (ms). More importantly, as the complexity of the problem graphs scale, with it scales the latency to solve the problem. With our method, we see a linear progression of time to solution as graph size increases rather than the previously seen exponential curve. 
% \input{sections/problemstatement}
%\input{sections/methodology}
\section{System Architecture and Design} 
\label{sec:method}
\begin{figure*}[htb]
\centering
\includegraphics[width=0.8\linewidth]{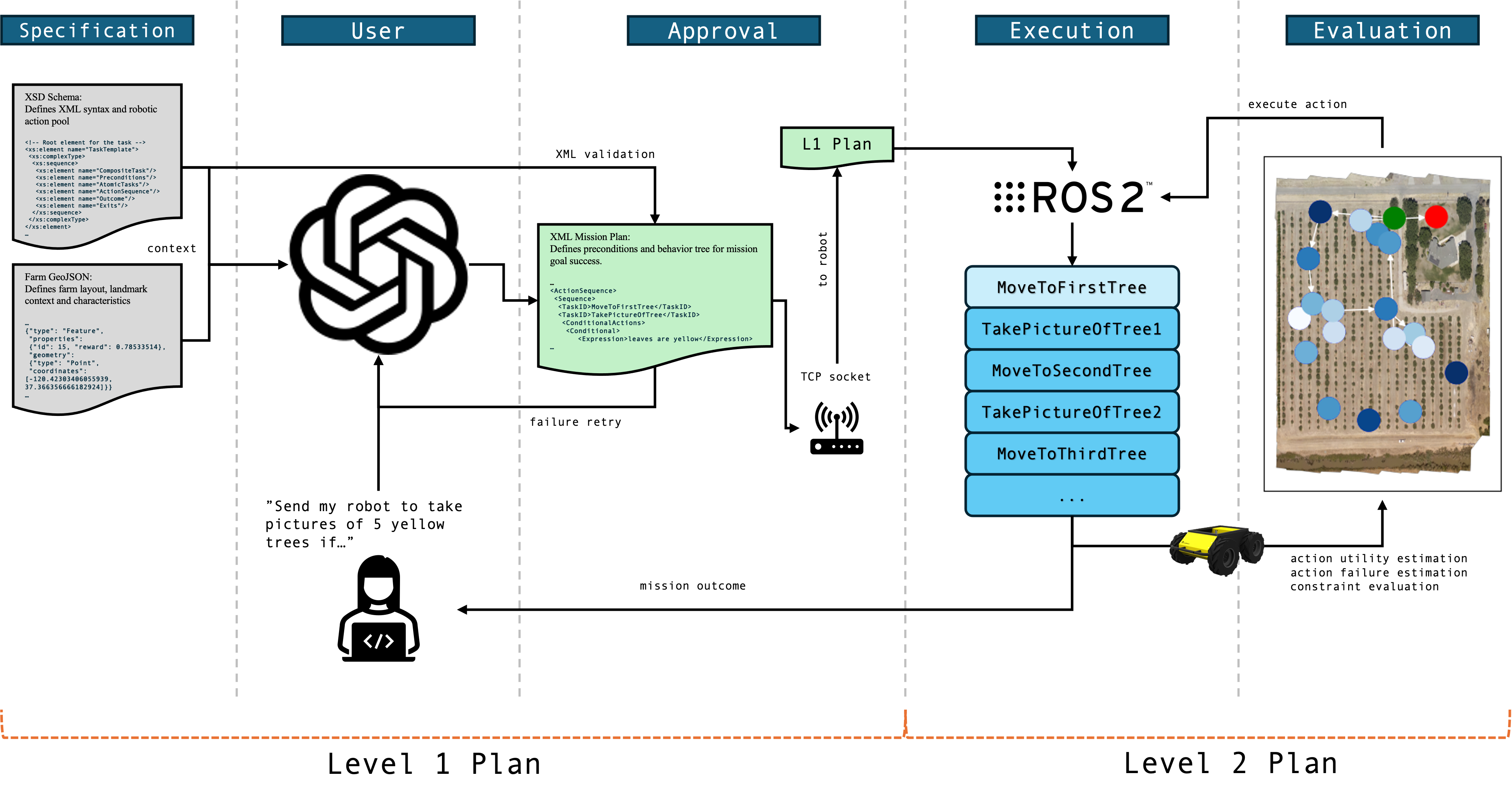}
\caption{Our proposed architecture follows the guidance of \cite{noauthor_ieee_2024} and breaks up MP into five functional roles. We place each of our software modules into at least one of the five
blocks to get a data pipeline that ingests context and natural language, creates an offline L1 plan, converts -- at run time -- said plan into a L2 plan, and then executes it. }
\label{fig:gnn-sop-architecture}
\vspace{-3mm}
\end{figure*}
As mentioned in the introduction, one of our design objectives is to align the MP output with the IEEE standard 1872.1-2024 for representing robotic tasks \cite{noauthor_ieee_2024}. 
This choice is motivated by the goal of developing tools that can be easily interfaced with other systems following the same standard. While we considered 
alternatives such as  IEEE 1872.2-2021 \cite{IEEE2021}, we decided on 1872.1-2024 due to its focus on task specific representation.
%Given that this architecture is robot agnostic, we focus on solving MP through task representation rather than robot state and environment interaction.
Accordingly, the framework behind our architecture is broken up into the functional software blocks defined in \cite{noauthor_ieee_2024}: specification, user, approval, execution, and evaluation (see
Figure  \ref{fig:gnn-sop-architecture}).
% specification, approval, initialization, execution, evaluation, completion and assessment. 
These five  modules are further organized into two levels. 
Defined in \cite{noauthor_ieee_2024}, a Level 1 (L1) plan is an abstract mission format representation with a modular decomposition of tasks. 
An L1 plan specifies \emph{what} should be done, but not \emph{how} it should be done.
In our implementation, L1 plans are represented by  XML files following an XSD schema encoding the task representation defined in the standard. 
A Level 2 (L2) plan defines how each of the tasks in an L1 plan can be executed on a specific robot platform.
% That is, a L2 plan defines how the the tasks specified in a L1 plan 
% will be executed on a specific robot platform.
% I HAVE MOVED THE DETAILED EXPLANATIONS TO THEIR RESPSECTIVE SECTIONS
% In our work, an L2 plan is executed by a ROS2 node in charge of decoding the L1 plan and converting it into tasks that are the sequenced into a behavior tree. Additionally, we introduced a Level 3 (L3) plan, not defined in the standard.
% The third level of implementation is an evaluator, post-L2, and explicitly considers and enforces
% constraints and feasibility of L2 plans in the real world where the robot operates. {\color{red} But then one could ask: why is this not part of L2, 
% since L2 targets a specific robot platform?}

Core to the proposed approach is our cloud-based user interface powered by OpenAI's GPT-4o  which generates the L1 mission plan. 
 In the current implementation, a simple text based interface is exposed for the user to input a mission using natural language. There are no special computational requirements for this module other than internet connectivity and an OpenAI API token.
The second part of the pipeline consists of a set of 
ROS2 nodes processing missions represented
as L1 plans, and decoding them into L2 plans and then executing them. 
Depending on the robot platform used to execute the mission, these
will target specific hardware modules. As explained in Section \ref{sec:results},
in our case the L2 module targets a ClearPath Husky with a self-developed
sensor suite appropriate for the agricultural monitoring tasks we are interested
in.
We next describe the role of each
of the five functional modules.

% These nodes run on a ClearPath Husky equipped with an ambient temperature sensor, thermal camera, and CO$_2$ sensor as usable peripherals. 

%These tasks are then handed off to an SOP solver {\color{red} cite or update} to, in real-time, determine the value, capability, and possibility of achieving each task. 
%Level 1 and 2 plans are defined in \cite{noauthor_ieee_2024}.
%The architecture sketched in 
% Figure \ref{fig:gnn-sop-architecture}, shows how each of these modules corresponds to \cite{noauthor_ieee_2024}.

% {\color{red}As we will explain in the next section, the importance of the level 1 XSD schema is paramount in being able to generate a usable mission plan.
% Prior to schema curation, syntax failures were much more prevalent in the plan generation step with respect to schema validation. 
% In general, however, this is preventable through the closed-loop validation loop prior to plan execution. 
% The beneficial aspect of this architecture is that it abstracts away executable code between level 1 and 2 schemas to prevent malicious or invalid code execution, saving on time or security. 
% In many current LLM based planners, executable code is used at both levels.
% More importantly -- and a limitation we will cover in this paper -- failures in functional validation were just as common. 
% % While syntactic validation is important in seamless MP data flow, semantic validation is equally as important. 
% \textit{If you're able to generate a valid plan, but it doesn't do what you wanted it to do, is it really valuable}?}

\subsubsection{Specification} \label{sec:spec}
The \emph{specification} step provides the context 
necessary for the LLM to solve the MP problem.
It is within this context that a mission described
in natural language  is later converted into an L1 plan encoded in XML.
Accordingly, the specification consists of two files.
The first is an XSD L1 schema representing both the mission plan format as well as an available atomic action pool for the robot(s) being commanded.
While in some literature PDDL is used to to represent this  knowledge, 
 we chose XML for its established presence in the software engineering industry.
%In fact, this part of the specification has broad applicability and can be shared among many different MP problems.
Notably, the first instance of the XSD framework was also created by ChatGPT after being provided with the standard in \cite{noauthor_ieee_2024}. Starting from the initial XSD file produced
by ChatGPT, we then added  the robot task pool and schema level detail to ensure  task sequences are encoded as behavior trees. 
These manual additions were necessary because they were not part of the standard.
Moreover,  the addition of the robot task pool does require domain knowledge of which 
capabilities a specific robot has.\footnote{\label{fn:website}The
complete XSD schema, supplementary material, and the code are available on \url{ucmercedrobotics.github.io/gpt-mp-ros2}.}
%Additionally, enabling the L1 schema to generate behavior tree task sequences required 
%required adding some additional structures that were not part of the standard
Accordingly, we shaped the relationship between XML tags such that when an L1 plan is generated its resulting task sequence takes the shape of a behavior tree. 
This creates a conformance to both the IEEE standard and behavior tree modeling, something that can be used generally by other applications seeking to integrate a mission planner.
The second part of the specification provides instead the 
spatial context describing the environment where the mission will be
executed. In our  domain, this represents
the spatial layout of the specific farm where the mission will be executed.
This  is presented in Geographical JavaScript Object Notation (GeoJSON). This file contains locations and characteristics about the farm,
e.g., the GPS location of trees,  as well as other characteristics, such as tree attributes (age, type, recorded yield, etc.) and more.
In a typical use of our system, a user (described in the next subsection) 
would start by typing a desired mission query, e.g., ``\textit{Send the robot to take pictures of yellow trees in the northern half of the farm.}'' GPT-4o translates this request into a mission plan being aware of both the high level 
capabilities of the robot (as per the XSD file) and the geographic
area where the robot should operate (as per the GeoJSON file).

% These two specification files are the only domain specific items in the architecture. 

% Equipped with the above context and an MP query, GPT-4o ultimately returns an MP as an XML behavior tree formed task listing, complete with global state information and parameters. 
% This step is a closed-loop system whereby failure to validate the XML against the XSD is requested to be fixed by GPT-4o until it validates, ensuring decoding success. 

% \begin{figure}
% \centering
% \includegraphics[width=1\linewidth]{figures/xmlmp.png}
% \caption{Action Sequence portion of mission plan and a single Atomic Task generated by ChatGPT. Note this example partially contains what, ultimately, is decoded as a behavior tree in the ROS2 mission plan decoder node.}
% \label{fig:example_mp}
% \end{figure}

\subsubsection{User}
The \emph{user} role encompasses any entity that utilizes the specification to 
generate an output. In our implementation, there are two user entities, i.e.,  
the human end user who types the mission request, and the ChatGPT front end 
that generates an L1 plan. In this context, ChatGPT serves as an extension of the 
end user, bridging the gap between the user's textual query and the context 
provided by the specification.

Both user entities bring complementary expertise relevant to the precision 
agriculture domain. The human, with farming experience, can formulate 
natural language queries aimed at gathering domain-specific data. ChatGPT, 
empowered by the background knowledge of the large language model 
and the provided context, synthesizes 
mission plans based on underlying robotic capabilities, ultimately achieving the 
desired outcome -- even if the human has limited knowledge of the robotic 
platform and its capabilities.
This approach abstracts from the complexities of MP by allowing ChatGPT to take on the role of curator to eliminate the need for the human to require much context at all, farm or robot related.

\subsubsection{Approval}
After an L1 plan is generated, the \textit{approval} step validates the L1 output against the XSD schema provided in the specification.
This ensures that the mission plans can be sent offline to the robot for execution. 
During this phase, there is a closed loop with the ChatGPT agent ensuring that whatever mission is generated is syntactically valid. 
This step is driven by the observation that ChatGPT sometimes generates XML files that do not fully comply with the XSD provided in the specification.
When this happens, the L1 plan is sent back to ChatGPT together with the error identified by the XSD validator and ChatGPT is asked to fix the error. 
In all instances we observed that a single iteration through this feedback loop is sufficient to fix any detected error. 
It shall be noted that this iteration, when executed, is fully automated and does not necessitate of any human intervention.
After this first approval step, the L1 plan is sent to the robot, where the same validation step occurs on target, to verify successful transmission.
% Some sections of the paper require further 
% elaboration. For instance, in Section III.3, the authors 
% state, “..., another validation step occurs on target, to 
% verify successful transmission,” but do not explain how 
% this second validation occurs. 
This step is critical in that after this stage there is no further syntactic validation and the robot will be subject to whatever mission plan has been generated.
After these two validation steps, starting from a simple mission prompt provided by the human end user, a valid L1 plan encoded in XML and compliant with the 1872.1-2024 standard and the XSD schema has reached the robot.
On the target robot, a set of ROS2 nodes then converts an approved L1 plan into an L2 plan
by parsing the  XML file and decoding it to generate a task sequence. 
More specifically, the ROS2 nodes ensure that the L1 input follows the structure of a behavior tree, with attributes that define both each individual task and the overall system state.
While this step could in principle also be performed off the robot, 
in our implementation it is executed on the robot, because verifying robot configuration and preconditions can be done directly while decomposing tasks.
After decoding is successful, a list of task objects is generated in ROS2's messaging framework via a service. 
The robot will first check all preconditions to ensure it is ready to accept this plan. Then, a list of tasks in the form of a behavior tree are parsed and sequenced for the next block. 
%After the initial L1 plan is translated into L2 with tasks  managed by the ROS2 nodes, the robot disconnects from the planner front-end and  commences
%execution.

%For example, one atomic task is to move the robot to a desired location
%specified as a GPS coordinate. 
%Defined in this atomic type, named a Parameter by \cite{noauthor_ieee_2024}, is the latitude and longitude of each point, among other characteristics.

% Solving for an instance of an SOP on a complete graph, we use a GNN trained to solve for SOPs. This comes from our previous work in {\color{red} cite} aiming to create a real-time algorithm for solving SOPs. We use the MP generated by our ChatGPT module to create the sequenced task list, but curate and optimize it through our GNN-SOP solver. At runtime, we reevaluate our mission plan at each action with respect to distance traveled. Starting from the first tasks at a single point, we query the GNN based on current state (node attributes) and remaining budget. Given that our cost might be different than is computed directly between two points, due to stochasticity, we predict our next action after every successive, completed action. Ultimately, we will end up with a list of actions from our original MP that leads us back to our goal state as formally defined in Eq.~\eqref{eq:MPSOP}. During and at the end of path execution, we evaluate total reward gained per task that we execute. Rewards are only gained if a location is visited and tasks are executed at that location. 

\subsubsection{Execution}
During \emph{execution}, the robot  performs the sequence of tasks 
derived from the L2 plan
with the objective of achieving the goal specified by the human.
At this stage, the execution becomes a classic
 closed loop system until successful completion of the mission. 
Using the same example query from Section \ref{sec:spec}, a decomposed atomic L2 task would be presented to the navigational ROS2 software stack, Nav2, and the robot would head north.  
Note that it is up to ChatGPT to extract
from the GeoJSON file a GPS coordinate corresponding to ``north'' and
making it available to Nav2 via the pipeline we discussed so far.
The handshake between execution and evaluation (discussed in the next 
subsection) is shared by Nav2 and our custom ROS2 task orchestration node. 
Both will manage task execution and evaluation to varying degrees. 
Given execution and evaluation are tightly coupled, we will decouple the data flow and discuss the former in the remainder of this section with the latter in the next.
For navigation specific tasks, integral to applications in robotics
for precision agriculture, Nav2 is leveraged to ensure accurate execution via localization  relying on the Kalman filter node provided by ROS2
augmented with RTK GPS.
In our example case, the request to take pictures in the northern half of 
the orchard will result in a task moving the robot to a 
 specific GPS point satisfying the geographic request.
 Nav2 handles the navigation, together with obstacle avoidance 
 also leveraging our formerly developed custom extension for 
 robust navigation in agricultural environments \cite{CARPINICRA2024D}.
 %in the north of the farm and request that our framework handle path generation and %obstacle avoidance leveraging our navigational hardware stack.
% The path navigator packaged with Nav2 will receive said tasks and perform the evaluation of success.
For data collection and sensor tasks, we leverage peripheral management nodes that orchestrates the onboard sensors set to collect data. 
Following again  the same example query from Section \ref{sec:spec}, this node 
detects  yellow leaves via a camera. 
For each peripheral onboard the robot, a ROS2 service is executed to handle 
requests as the tasks come in.

\subsubsection{Evaluation}
% The role of evaluation in the context of standard 1872-1.2024 is important for several reasons. First, the open problems our lab focuses on are often stochastic. Therefore, there is a need for architectures with feedback loops. Second, we operate in a real-time space. Not just the notion of evaluation is important, but how quickly it is accomplished. Lastly, our problems are often constrained. While we aim to maximize or minimize some objective function, we often have to do so with some global constraints on our problem. 
As the robot executes  L2 tasks, \emph{evaluation} is implemented
on two different levels.
First, we  leverage ROS2's Nav2 stack  for handling dynamic 
evaluation of an initial path between two points.  
The path navigator packaged with Nav2 will receive said points and perform the evaluation of success via the same localization techniques that helped it traverse the path.
In the second level, we incorporate custom ROS2 nodes that query data collection for use in adhering to constraints or abiding by the behavior tree.
In our example, the robot would constantly be checking navigational status while executing movement between two points: the start point and a tree in the north of the farm.
Upon reaching the tree, the next sequential task, taking a picture, would be sent to the picture action server and the orchestrator then waits for a response on whether the tree is  yellow or not.
In the experiments  presented in the next section,
we show how evaluation also tackles some tasks that the LLMs alone 
cannot properly solve.
%In the first instance, we will show conditionally complex task sequences that require runtime evaluation of task outcomes. 
In our application domain we have
to assume an offline scenario where there is no network connectivity,
thus preventing ChatGPT from possibly revising a mission plan
that is not progressing as expected.
For example, if the user specifies a mission with resources 
constraints, the evaluation module monitors progress 
and intervenes if run time execution shows that the constraints will likely
be. This is particular important because we observed
ChatGPT's limitations in solving  optimization tasks
and its occasional inability to properly reason about space.
Additionally, the user may specify a mission without constraints
because it ignores the the limitations of the robotic system 
that will execute the mission.
 For example,
robots operating in orchards have limited energy budget and must
return to the charging station or deploy point before their batteries
are depleted. To complicate matters, the consumed energy is not 
fully predictable upfront, but is rather a random variable.
This problem is known to be an instance of 
the  the stochastic orienteering problem (SOP), 
a problem we have formerly studied in the context of 
precision agriculture \cite{9636669,CarpinTASE2024}
With a constraint on total distance traveled and no way of pre-planning,
due to the stochastic nature of the travel, the ability
to manage these constraints at run time is delegated to the
evaluation step.
It is our contention that the proposed architecture not only solves for SOPs, but can be generalized to solve for other optimization
problems that cannot be fully solved offline by the LLM.
We will show the results for solving SOPs in Section \ref{sec:SOP}.
\section{Results} \label{sec:results}

In this section we show how the system we described can synthesize and execute complex
missions starting from high level descriptions  in natural language. Due to space
limitations, only a subset of results is presented. The companion video, as well as the website linked in Section \ref{sec:spec}, provide additional details and visuals for the interested reader, as well as the code.

\subsection {Experimental Setup}
All missions are run on a ClearPath Husky controlled by an Nvidia Jetson Orin Nano
running ROS2 Humble. The robot is  equipped with a U-blox ZED-F9P GNSS module and with a Bosch BMI085
IMU for localization (both feeding the extended Kalman Filter node
part of ROS2 and used for outdoor localization).
We used GPT-4o-2024-05-13 with a temperature of $0.2$ and  set  max response tokens to 4096, i.e., the max supported by GPT-4o. While 
this could be a limitation in that  complex mission plans could be longer than 4096 tokens,  in our experiments  this never happened.
%With this in mind, we organize our experimentation as follows.

We assess ChatGPT’s ability to solve MP problems in precision agriculture, focusing on generating complex conditional mission plans using behavior trees and evaluating mission success. 
We compare its performance on non-spatial and spatial planning problems of varying complexity.
Building on these findings, we demonstrate the need for additional planning software to optimize LLM-based MP. 
Using our Husky robot in a real-world testbed, we showcase mission execution with and without supplementary planning tools.
Finally, we evaluate how easily ChatGPT's output can be generalized across different problem types. 
Full mission prompts are available on the website and video, with Table \ref{table:results} showing abbreviated versions.
Our study aims to answer:
1. \textit{Can we use ChatGPT to solve offline MP problems without any supporting planning software modules?} 
2. \textit{Can ChatGPT understand spatially complex mission queries?} 
3. \textit{Can we leverage the generality of LLMs to optimize mission planning problems with constraints?}
% Some sections of the paper are overly lengthy (Such as 
% Section IV.A second paragraph). Reducing the length of 
% these sections would create more space for meaningful 
% content and enhance overall clarity. 
% ORIGINAL TEXT BELOW
% First we  assess the ability of  ChatGPT to solve MP problems in precision agriculture. We will show how the LLM can generate complex conditional mission plans through the use of behavior trees and the rate of success. We will also show the difference in solving non-spatial and spatial planning problems of various complexities. 
% Based on those initial findings, we will demonstrate a need for additional supportive planning software to optimize LLM based MP. Carried out by our Husky at our testbed, we will demonstrate solving a real-world precision agriculture problem with and without additional planning software.
% Last, we will assess how simply we can generalize ChatGPT output to solve more than one type of problem. 
% All of these experiments originate from mission prompts that can be found in full on the website and in the video
% while Table \ref{table:results} shows shortened versions of the full prompts,
% which do not fit in the table.
% With these experiments we aim to answer the following questions: 
% 1. \textit{Can we use ChatGPT to solve offline MP problems without any supporting planning software modules?} 
% 2. \textit{Can ChatGPT understand spatially complex queries?} 
% 3. \textit{Can we leverage the generality of LLMs to include optimization and constraints into mission planning problems?}

\subsection{Solving MP problems and Spatial Awareness} \label{sec:bh}
\begin{figure}[htb]
\centering
\includegraphics[width=1\linewidth]{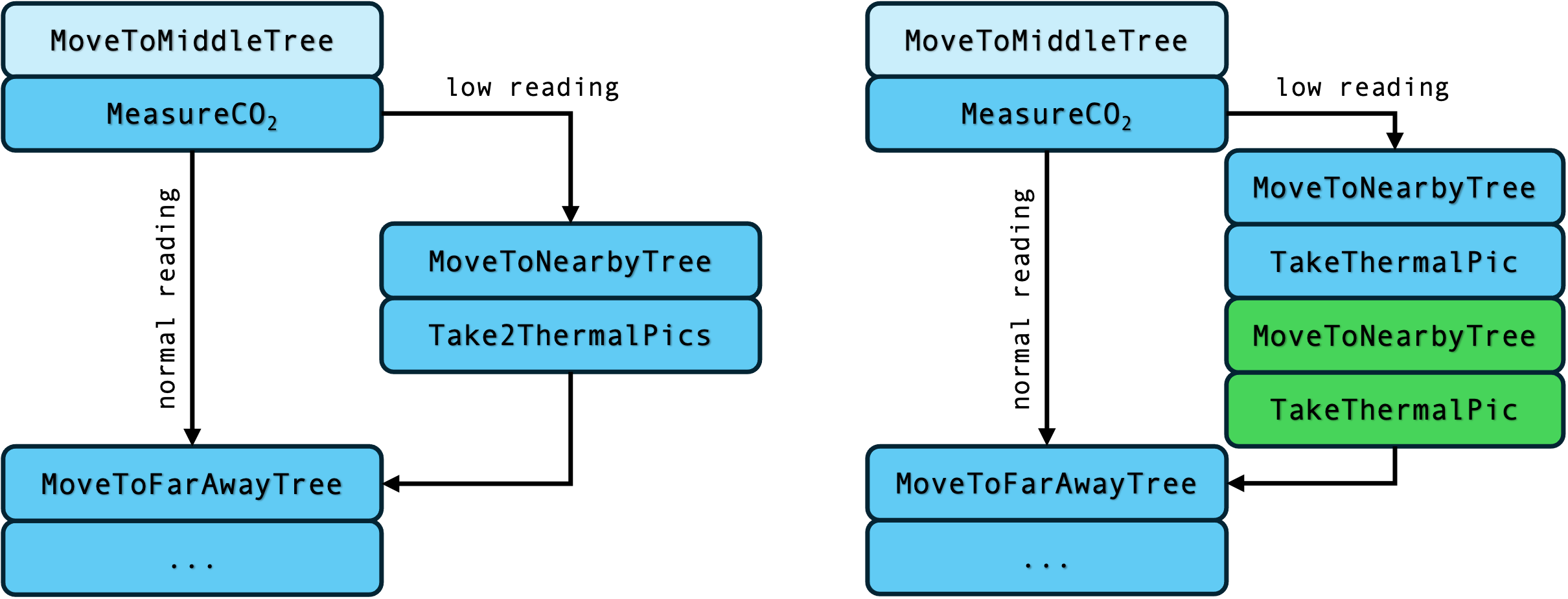}
\caption{Demonstration that without explicit intent, sometimes nuanced statements are missed. Here, we see two different behavior trees from two very similar queries. The added tasks are highlighted in green with the starting tasks in light blue.}
\label{fig:explict_query}
\end{figure}

We began by evaluating ChatGPT's capability in solving general MP problems
with and without spatial awareness requirements. Similar to previous works \cite{mower_ros-llm_2024, kannan_smart-llm_2024, macaluso_toward_2024}, our goal is to assess ChatGPT's proficiency in generating behavior trees that support mission specifications involving navigation and
data acquisition tasks.
%It shall be noted that, as anticipated in Section \ref{sec:method}, the XSD 
%initially produced by ChatGPT, by analyzing the IEEE standard, did not explicitly included
%structures to handle behavior trees, and those had to be added by hand.
%After addressing our schema problems, 
To this end, we 
 asked an array of mission queries to understand ChatGPT's limitations with respect to conditional complexity and spatial awareness.  
% We developed a set of mission queries with the intent of exploring a wide range of missions that might be typically asked to an autonomous robot. 
The set of queries range from simple, complex, conditional, to spatial. 
Note that all missions were manually reviewed for semantic success. Their results are found in Table \ref{table:results}.
Syntactically, all missions generated from ChatGPT were correct and paths generated were feasible.
\begin{table}[t]
\begin{center}
    \begin{tabular}{cccc}
    \cline{1-4}
     \multicolumn{1}{c}{Mission Queries} & \multicolumn{1}{c}{Number of Tasks} & \multicolumn{1}{c}{Success?}\\\cline{1-4}
     % \multicolumn{1}{c|}{} & $R$ & $F$ & $R$ & $F$ & $R$ & $F$ \\\cline{2-7}
     \hline
      \textbf{non-spatial reasoning}&&& \\
     \hline
     "\textit{3 pictures in row of 5}" & 8 (0 conditionals)& True \\ 
     "\textit{4 trees, 2 sensors}" &8 (0 conditionals)& True \\
     "\textit{Reward shaping}" &20 (0 conditionals)& True \\
     "\textit{5 nested if conditionals}" & 14 (5 conditionals) & True\\
     "\textit{If-else with nesting}" & 16 (5 conditionals) & True\\
     \hline
     \textbf{spatial reasoning}&&& \\
     \hline
     "\textit{4 corners relative}" & 8 (0 conditionals) & False\\
     "\textit{4 corners absolute}" & 8 (0 conditionals) & False\\
     "\textit{Sample 100 meters of trees}" & 10 (0 conditionals) & True$^*$\\
     "\textit{North, center, east samples}" & 6 (0 conditionals) & False\\
     "\textit{Relative conditionals}" & 8 (2 conditionals) & False\\
     "\textit{Relative + absolute conditionals}" & 17 (2 conditionals) & False\\     
     \hline
    %  \hline
    %  $\text{graph50}_{B=2^*}$ & 0.1 & 6.952 & 0.533 & 18\%$^!$ & 6.422 & 45.935 & 10\%\\  
    %  $\text{graph60}_{B=2^*}$ & 0.1 & 8.889 & 0.756 & 30\%$^!$ & 7.429 & 148.93 & 11\% \\    
    %  $\text{graph70}_{B=2^*}$ & 0.1 & 8.462 & 0.778 & 40\%$^!$ & 8.381 & 205.451 & 12\%\\   
    % \hline
    \end{tabular}
    \caption{
    Assessment of various mission prompts presented to our MP system. $^*$ technically correct, but extremely suboptimal. See Section \ref{sec:SOP} for more.
    % "Number of Tasks": total atomic tasks, including conditional, decomposed. "Success": semantically represents the prompt. All missions were syntactically valid.
    }
    \label{table:results}
\end{center}
\vspace{-8mm}
\end{table}
When dealing with all forms of non-spatial MP (i.e., requests not asking to reason about space), 
our architecture never failed. 
From simple tasks like taking three pictures in a row of five trees to more complex conditional tasks like taking samples of trees depending on the outcomes of previous samples, ChatGPT proved to generate semantically valid and feasible mission plans. However, 
while we were able to generate behavior tree logic for complex conditional navigational missions, 
doing so  sometimes required  an extremely detailed specification in natural language -- something 
a non-specialist user would not easily produce.
% We did however encounter an instance in which ChatGPT required more specific language in the query to properly generate a semantically correct mission. 
More precisely, an example request was, ``\textit{From the starting point, find a tree somewhere in the middle of the orchard and measure $CO_2$. If low reading, go to and take 2 pictures of nearby trees. From there, measure another tree as far away as possible for $CO_2$ and repeat the same process. Once done, take a temperature reading at any one of the northern most trees. Finally, return to end}." 
The problem comes from the second sentence, where we had originally intended that the robot visits two unique trees nearby, with ``trees'' being plural, and take a single picture at each of them. 
However, no matter the configuration, the LLM produced a mission plan to visit a single tree and 
take two pictures of it. 
After changing the second sentence to, "\textit{If low reading, go to 2 different trees nearby and take a picture of each of them}" 
we finally saw the intended output. The nature of this problem is 
visualized in Figure \ref{fig:explict_query}.
In requiring this explicit distinction, we see the need for effective functional validation practices shown by \cite{jansen_can_2023}
and the venue for future work.

\begin{figure}[htb]
\centering
\includegraphics[width=1\linewidth]{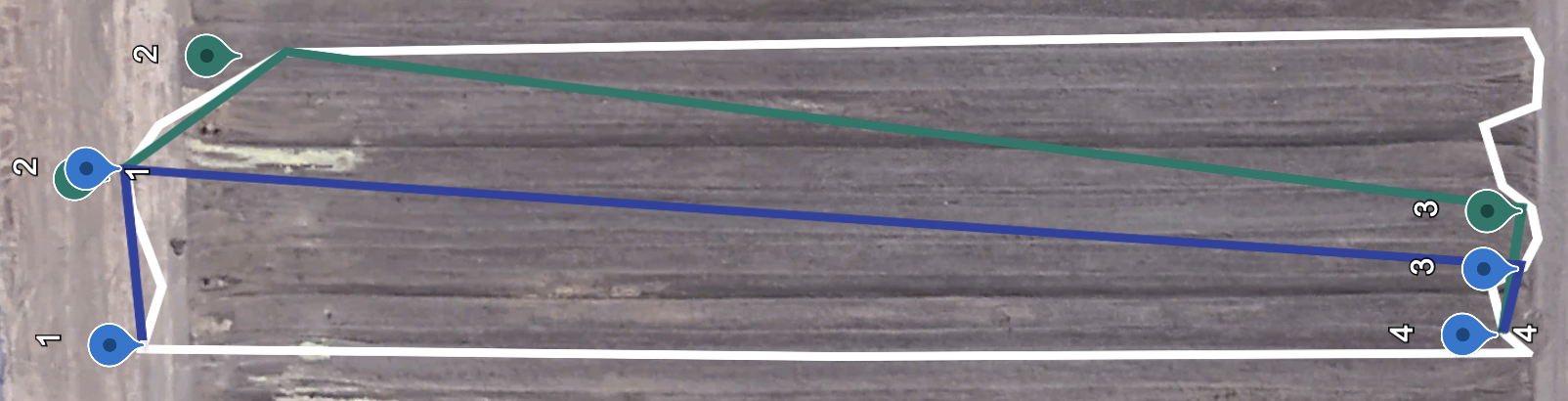}
\caption{With left being north, we asked ChatGPT to trace the 4 corners of our "rectangular" farm layout, shown in white. The green is a relative query (corners of rectangle) and the blue is an absolute query (cardinal direction).}
\label{fig:spatial}
\vspace{-5mm}
\end{figure}

Next, we tested our system on more complex, realistic missions that would come from a typical data aggregation expedition. 
This  often involves some form of spatial awareness or reasoning, and we correspondingly
broke this assessment into two parts: absolute and relative spatial cognition.
In these cases we tested ChatGPT's ability to interpret prompt including terms such as ``north'' or ``south'' and simple spatial concepts like what is the shape of the orchard in which the robot operates.
During this testing, as Table \ref{table:results} suggests, we almost immediately ran into problems. While all of the missions were 
syntactically valid, ChatGPT often times misunderstood the geometry of the farm or its orientation. 
In two of the queries, "\textit{4 corners absolute/relative}" in Table \ref{table:results}, we asked our LLM planner to trace the four corners of our farm plot. 
As seen in Figure \ref{fig:spatial}, ChatGPT creates a  path through the orchard which is close but not actually what we asked for. 
A similar output in "\textit{Sample 100 meters of trees}", where we ask ChatGPT to sample the most trees possible in 100 meters and it only samples roughly  trees spanning 17 meters which is is obviously suboptimal.
These nuances demonstrate the limitations of using only ChatGPT output to guide a navigational stack. In the next subsection, we will expand on this last mission prompt and show how adding a spatially aware software module
as part of the evaluation block can immensely improve performance.

\subsection{Optimization and constraints} \label{sec:SOP}
As we just saw, the current version of ChatGPT 
does not display the type of spatial awareness needed for MP in 
precision agriculture \cite{li_advancing_2024}. 
Therefore, given that many of the problems we study have an inherent 
optimization component related to spatial awareness, 
we included in our system a spatial planning module to overcome ChatGPT's current limitations. 
\begin{figure}
\centering
\includegraphics[width=0.5\linewidth]{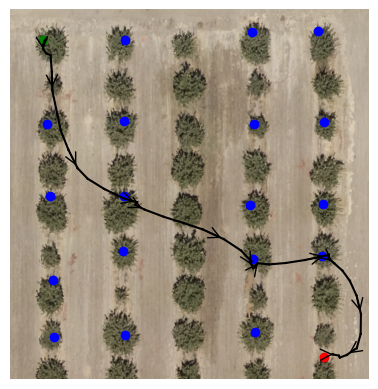}
\caption{A plot of our Husky robot's GPS path on our pistachio orchard testbed while solving for an instance of a mission plan where not all trees
can be visited with the allocated budget.}
\label{fig:path}
\end{figure}
For these experiments, we follow the architecture drawn in Figure \ref{fig:gnn-sop-architecture} and add to the validation block a stochastic orienteering solver \cite{zuzuarregui_solving_2024}  to enable constrained route
optimization.
We focused on stochastic orienteering -- a variant of deterministic orienteering particularly relevant in precision agriculture-- in which an agent must  maximize the utility collected by visiting a set
of locations while subject to a hard bound on the traveled distance
due to the limited energy provided by the onboard battery.
Uniquely, distance traveled is not known until after path execution as the environment is, as the name suggests, stochastic.
Performance is measured in two ways: collected utility $R$, and failure, $F$, defined as the fraction of missions that violate the
constraints.
We use these metrics to compare the offline mission provided by ChatGPT
used alone or in conjunction with an optimization algorithm (called GPT-SOP
in the following) as part of the evaluation block. 
% For completeness
% we compare also against another offline optimal method based
% on mixed integer linear programming \cite{MILP} that is however extremely time
% consuming (referred to as MILP in the following).
For completeness, we also show the results with the optimal offline mixed integer linear programming (MILP) \cite{MILP} solver that, for graphs above size 20, must be capped at 10 minutes per solution, while GPT alone operates in the seconds range and \cite{zuzuarregui_solving_2024} in the milliseconds.
% Another 
% concern is the authors say that MILP, which has competitive 
% performance and outperforms the presented methods with 
% regards to utility, takes an "extremely long time". This 
% computational time difference should be quantified to give 
% the reader a full understanding of the performance 
% trade-offs between the proposed method and the baseline. 
This  gives us measurable metrics showing how our SOP online solver is 
essential to the architecture. 
To grasp the significance of this addition, we started by 
asking ChatGPT to collect pictures of a large set of trees in an orchard, 
ignoring the fact that the robot does not have sufficient 
autonomy to visit them all. 
Figure \ref{fig:path} shows that even though ChatGPT produces
an L1 plan aiming at visiting numerous trees (marked in blue)
the GPT-SOP component embedded in the evaluation plan intervenes
and limits the visit to a few, ensuring the robot reaches 
the end point (red dot) before it runs out of energy.

\begin{table}[t]
\begin{center}
    \begin{tabular}{||c|c|c|c|c|c|c||}
    \cline{2-7}
     \multicolumn{1}{c|}{} & \multicolumn{2}{c|}{GPT-SOP Solver} & \multicolumn{2}{c|}{ChatGPT} & \multicolumn{2}{c||}{MILP}\\\cline{2-7}
     \multicolumn{1}{c|}{} & $R$ & $F$ & $R$ & $F$ & $R$ & $F$ \\\cline{2-7}
     \hline
     $\text{graph20}_{B=2}$ & 2.092 & 11\% & 2.095 & 4\% & 3.414 & 12\%\\
     $\text{graph30}_{B=2}$ & 5.592 & 9\% & 2.531 & 10\% & 6.973 & 10\%\\
     $\text{graph40}_{B=2}$ & 5.785 & 7\% & 2.710 & 13\% & 8.843 & 10\%\\         
     \hline
    %  \hline
    %  $\text{graph50}_{B=2^*}$ & 0.1 & 6.952 & 0.533 & 18\%$^!$ & 6.422 & 45.935 & 10\%\\  
    %  $\text{graph60}_{B=2^*}$ & 0.1 & 8.889 & 0.756 & 30\%$^!$ & 7.429 & 148.93 & 11\% \\    
    %  $\text{graph70}_{B=2^*}$ & 0.1 & 8.462 & 0.778 & 40\%$^!$ & 8.381 & 205.451 & 12\%\\   
    % \hline
    \end{tabular}
    \caption{
    Overall results from benchmarks used in \cite{9926510} averaged over 100 trials. 
    }
    \label{table:sop-results}
\end{center}
\vspace{-5mm}
\end{table}

In Table \ref{table:sop-results}, we show numerical results where Chat-GPT is asked to produce plans visiting as many trees as possible on orchards with different number of trees (varying from 20 to 40) while sticking to a limit on the maximum traveled distance (parameter $B$).
Note, $B$ has no units due to being based on the normalized total size of the graph in any units.
% Lastly, there are a few issues with the results presented in Table II. First, it is unclear what units B is in. 
As we can see, ChatGPT alone (middle), produces plans that are extremely conservative in terms of collected rewards $R$, while still
incurring in relatively high failure rates. 
Once the system is augmented with our online GPT-SOP solver (left), results dramatically improve.
When reading the rationale ChatGPT sends with the mission plan, it expresses that it is using some form of a greedy strategy when selecting nodes. 
Due to the complexity presented in SOPs, such a simple heuristic will only perform so well. 
Plans generated by GPT alone will succumb to more failures due to being offline solutions,
whereas \cite{zuzuarregui_solving_2024} implements an online solution that is trained to understand stochasticity.
\section{Conclusions and Future Work} \label{sec:conclusions}
In this paper we presented a full pipeline that leveraging LLMs starts from a high level
objective specified in natural language and produces mission plans that can be executed
by a robot operating in an orchard with no internet connectivity. To the best of
our knowledge, this is the first example leveraging LLMs in this domain.
The solution we proposed provides a valuable tool to enable non specialists
to utilize robotics and AI in precision agriculture.
Our experimentation has shown that LLM can be beneficial, but may fall short
when facing tasks requiring spatial awareness or constrained optimization.
These limitations are overcome by augmenting the system with 
optimization algorithms part of the evaluation block in the proposed 
architecture. Future research will explore how this architecture
can be extended to systems featuring multiple robots, as well as
robots interacting with fixed deployed infrastructure such as static
sensors.

\begin{comment}
user to mission plan intended to control autonomous robots in precision agricultural scenarios. 
Specifically, when aggregating sensor data at scale, the problem becomes a SOP. 
We show a competitive solution to this problem as a part of the architecture. 
This data flow goes from a cloud-based LLM, ChatGPT, level 1 mission planner, to ROS2 based level 2 task decoder, to Tensorflow based GNN state evaluator. 
We also showed that we can standardize, with the help of \cite{noauthor_ieee_2024}, the method in which tasks are defined, sequenced, and executed. 
Finally, our experiments delved into the limitations of using ChatGPT as a mission planner with limited domain specific knowledge and a generalized approach to using ChatGPT as a planner.

In the future, we wish to explore the avenue of formal software verification with this system architecture as in \cite{jansen_can_2023}. 
We showed that while we can generate feasible mission plans, feasible doesn't necessarily equate to formally verified. 
With an existing architecture, formalizing the mission plan can give us a quantitative assessment as to how good of a mission planner LLMs can be, instead of an informal or human-in-the-loop verification. 
We can do so with linear temporal logic (LTL) or any alternative. 

\end{comment}    
%\input{sections/conclusions}  

\bibliographystyle{plain}
\bibliography{report.bib}

\end{document}